\documentclass[a4paper, 10 pt, conference]{ieeeconf}  

\IEEEoverridecommandlockouts                              

\usepackage{graphics} 
\usepackage{epsfig} 
\usepackage{mathptmx} 
\usepackage{times} 
\usepackage{amsmath} 
\usepackage{amssymb}  
\usepackage{algorithm}
\usepackage{algpseudocode}

\usepackage{cite}
\usepackage{amsmath,amssymb,amsfonts}
\usepackage{graphicx}
\usepackage{textcomp}
\usepackage{xcolor}
\usepackage{comment}
\usepackage{ulem}
\usepackage{multirow}
\usepackage{url}

\title{\LARGE \bf
Monte Carlo Beam Search for Actor-Critic Reinforcement Learning in Continuous Control}

\author{Hazim Alzorgan$^{1}$ and Abolfazl Razi$^{2}$
\thanks{*This work is supported by the National Science Foundation under Grant Number 2204721 and MIT Lincoln Lab under Grant Number 7000565788.}
\thanks{$^{1}$Hazim Alzorgan is a PhD student, School of Computing,
        Clemson University, Clemson, SC 26935, USA
        {\tt\small halzorg@clemson.edu}}%
\thanks{$^{2}$Abolfazl Razi is an Associate Professor and the Director of AI-based Sensing, Networking, and Data Services (AI-SENDS) Laboratory, School of Computing,
        Clemson University, Clemson, SC 26935, USA
        {\tt\small arazi@clemson.edu}}%
}

\begin{document}

\maketitle
\thispagestyle{empty}
\pagestyle{empty}

\begin{abstract}
 Actor-critic methods, like Twin Delayed Deep Deterministic Policy Gradient (TD3), depend on basic noise-based exploration, which can result in less than optimal policy convergence. In this study, we introduce Monte Carlo Beam Search (MCBS), a new hybrid method that combines beam search and Monte Carlo rollouts with TD3 to improve exploration and action selection. MCBS produces several candidate actions around the policy’s output and assesses them through short-horizon rollouts, enabling the agent to make better-informed choices. We test MCBS across various continuous-control benchmarks, including \texttt{HalfCheetah-v4}, \texttt{Walker2d-v5}, and \texttt{Swimmer-v5}, showing enhanced sample efficiency and performance compared to standard TD3 and other baseline methods like SAC, PPO, and A2C. Our findings emphasize MCBS’s capability to enhance policy learning through structured look-ahead search while ensuring computational efficiency. Additionally, we offer a detailed analysis of crucial hyperparameters, such as beam width and rollout depth, and explore adaptive strategies to optimize MCBS for complex control tasks. Our method shows a higher convergence rate across different environments compared to TD3, SAC, PPO, and A2C. For instance, we achieved 90\% of the maximum achievable reward within around 200 thousand timesteps compared to 400 thousand timesteps for the second-best method.
\end{abstract}


\section{Introduction}\label{sec:intro}

Reinforcement Learning (RL) has made significant strides in recent years, achieving success in various fields such as robotics, autonomous driving, and gaming \cite{rl1, rl2, rl3, pedram}. Traditional RL methods often concentrate on discrete action spaces or impose strict assumptions about the environment, which can limit their effectiveness in real-world situations. However, continuous control tasks—where the action space is both continuous and high-dimensional—are becoming increasingly important in practical applications like robotic manipulation \cite{robot1}, continuous locomotion \cite{mujoco1}, and industrial control \cite{control1}. As these tasks become more complex, the challenge of designing algorithms that can effectively learn optimal control policies persists.

A commonly used framework in continuous-action RL is the actor-critic family of algorithms, which separates the policy (actor) from the value estimation (critic). Specifically, the Deterministic Policy Gradient (DPG) framework served as the basis for Deep Deterministic Policy Gradient (DDPG) \cite{ddpgpaper}, allowing for end-to-end training of deep networks in high-dimensional continuous control settings. Despite its achievements, DDPG faces issues such as overestimation bias, sensitivity to hyperparameters, and training instability. Twin Delayed Deep Deterministic Policy Gradient (TD3) \cite{td3paper} tackles these challenges by implementing three essential techniques: \textbf{Clipped Double Q-learning} to mitigate overestimation bias, \textbf{Delayed Policy Updates} to enhance learning stability, and \textbf{Target Policy Smoothing} to improve robustness. These adjustments significantly boost TD3’s stability and sample efficiency compared to DDPG across various continuous control benchmarks \cite{td3paper, bench1, pedram2}.

Despite these advancements, exploration continues to pose a significant challenge in continuous-action RL. Standard methods usually add Gaussian or Ornstein-Uhlenbeck noise to the actor’s output \cite{ddpgpaper}, which encourages local exploration but often falls short in tasks that require strategic long-term planning. To overcome this limitation, look-ahead methods that evaluate multiple potential actions can improve exploration and refine policies. Specifically, beam search and Monte Carlo Tree Search (MCTS) have shown great effectiveness in discrete-action environments, such as game-playing agents like AlphaGo and AlphaZero \cite{alphago, alphazero}. However, adapting these methods to continuous control is complex due to the vast and unstructured nature of the action space.

Beam search systematically expands a set of candidate actions, known as a “beam,” and prunes them based on an evaluation function \cite{beamsearch1, beamsearch2}. Although it is primarily utilized in natural language processing, beam search can also serve as a local action-refinement tool when combined with a learned Q-function. By generating several action candidates around the policy’s output and selecting the best one based on the critic’s estimated Q-values, the agent can effectively carry out short-horizon look-ahead optimization.

MCTS is a planning algorithm that has seen significant success in environments with discrete action spaces. It estimates action values by conducting numerous rollouts (simulations) from the current state \cite{mcts1, mcts2}. However, using standard MCTS for continuous control necessitates specialized techniques like progressive widening to manage the infinite action space. Additionally, keeping a structured search tree in high-dimensional spaces can lead to considerable computational costs.

In this paper, we present a new hybrid approach that combines TD3 with a Monte Carlo Beam Search method for continuous-action environments. Unlike traditional TD3, which chooses actions based on single-step exploratory noise, our approach assesses multiple candidate actions—generated through beam search or short-horizon rollouts—to enhance the action selection process. Specifically, we introduce:

\begin{enumerate}
    \item \textbf{Beam Search for Candidate Generation:} A set of $B$ candidate actions is sampled around the policy’s output using Gaussian noise, forming a \emph{beam} of potential control actions.
    \item \textbf{Monte Carlo Rollouts (MCTS-like Evaluation):} For each candidate action, we perform short-horizon rollouts using a forward model or simulator, accumulating rewards and bootstrapping with the learned critic at the final state.
    \item \textbf{Action Selection via Critic Evaluation:} The agent selects the action that maximizes the estimated return, facilitating more informed exploration and policy updates.
\end{enumerate}

Our key hypothesis is that even a small beam width and a limited number of simulated steps can greatly enhance the agent's ability to avoid local optima and improve its policy more effectively than simple noisy exploration. By utilizing the critic's learned insights and simulating short-horizon trajectories, the agent can predict high-reward actions without needing extensive or full-depth tree expansions. We refer to this method as Monte Carlo Beam Search (\textbf{MCBS}) in the context of TD3.

\subsection{Contributions}

The primary contributions of this work are as follows.

\begin{itemize}
    \item \textbf{Novel Hybrid Algorithm:} We introduce MCBS, a framework that leverages beam search with short-horizon MCTS rollouts within the TD3 algorithm, tailored for continuous control tasks.
    \item \textbf{Implementation and Analysis:} We offer a thorough implementation of MCBS, looking into its computational cost and trade-offs. We explore important design decisions, such as beam width, rollout depth, and noise clipping, to evaluate how they affect performance.
    \item \textbf{Empirical Evaluation:} We carry out thorough experiments on established continuous control benchmarks, showing that MCBS enhances sample exploration and efficiency compared to TD3. Furthermore, we conduct ablation studies to examine the impacts of beam width and depth in MCTS simulations.

\end{itemize}

\section{Background and Related Work}\label{sec:background}

\subsection{Reinforcement Learning Formulation}

Reinforcement Learning considers an agent interacting with an environment in discrete time steps. Let $\mathcal{S}$ denote the state space and $\mathcal{A}$ the action space, which can be continuous. At each time step $t$, the agent observes a state $s_t \in \mathcal{S}$ and selects an action $a_t \in \mathcal{A}$ according to a policy $\pi$. The environment then transitions to a new state $s_{t+1}$ and provides a scalar reward $r_t$, following the transition dynamics $\mathcal{P}(s_{t+1} \mid s_t, a_t)$. The objective in RL is to learn a policy $\pi$ that maximizes the cumulative discounted return:

\begin{equation}
    J(\pi) = \mathbb{E}_{\tau \sim \pi} \left[\sum_{t=0}^{\infty} \gamma^t r_t\right]
\end{equation}

where $\gamma \in (0,1)$ is the discount factor, and the expectation is taken over the trajectories $\tau = \{(s_t, a_t)\}$ generated by $\pi$ under the environment dynamics $\mathcal{P}$.

For continuous action spaces, policy gradient methods commonly parameterize the policy $\pi_\theta(a \mid s)$ using neural networks with parameters $\theta$. Actor-critic methods decompose the learning process into two components:

\begin{enumerate}
    \item \textbf{Actor:} A policy network $\pi_\theta$ that produces actions (or action distributions) given states.
    \item \textbf{Critic:} A value function (or Q-function) $Q_\phi(s, a)$ that estimates the expected return for taking action $a$ in state $s$ and following $\pi_\theta$ afterward.
\end{enumerate}

This framework demonstrates strong empirical performance in tasks with high-dimensional or continuous action spaces \cite{rl1, rl2}.

\subsection{Actor-Critic Methods}

\subsubsection{Deterministic Policy Gradient (DPG)}
DPG methods parametrize the policy as a deterministic function $\mu_\theta(s)$ instead of a stochastic one. The policy gradient is estimated by backpropagating through the critic.

\begin{equation}
    \nabla_\theta J(\theta) \approx \mathbb{E}_{s \sim \mathcal{D}} \Big[\nabla_a Q_\phi(s, a) \big\rvert_{a=\mu_\theta(s)} \nabla_\theta \mu_\theta(s)\Big]
\end{equation}

where $\mathcal{D}$ is a replay buffer that stores past experiences.

\subsubsection{Twin Delayed Deep Deterministic Policy Gradient (TD3)}

Although DDPG has been widely adopted, it faces issues with overestimation bias in the Q-function, which in turn leads to unstable training. TD3 \cite{td3paper} mitigates these issues through three key improvements:

\begin{itemize}
    \item \textbf{Clipped Double Q-learning:} TD3 maintains two independent critic networks, $Q_{\phi_1}$ and $Q_{\phi_2}$, which are trained simultaneously. The target Q-value for the next state-action pair $\bigl(s_{t+1}, \mu_{\theta'}(s_{t+1})\bigr)$ is computed by taking the \emph{minimum} of $Q_{\phi_1}$ and $Q_{\phi_2}$ to reduce overestimation bias.
    \item \textbf{Delayed Policy Updates:} Unlike DDPG, TD3 updates the policy (actor) and target networks less frequently than the critics. Generally, the critic is updated at each time step, whereas the actor is updated only every $d$ steps (commonly $d=2$). This approach stabilizes training by making sure the critic is accurate enough before any changes are made to the policy.
    \item \textbf{Target Policy Smoothing:} To prevent the policy from exploiting spurious peaks in the Q-function, TD3 injects small clipped noise into the target policy action when computing the target Q-value.
\end{itemize}

Such enhancements greatly improve the stability and sample efficiency of TD3 when compared to DDPG. However, even with these advancements, TD3—similar to many policy gradient algorithms—depends on action noise for exploration. This approach may not be ideal in environments that necessitate more structured exploration strategies. In this context, we investigate alternative search-based methods designed to enhance exploration in TD3.

\subsection{MCTS}

\subsubsection{Overview of MCTS in Discrete Domains}

MCTS has emerged as a leading strategy for making decisions in large discrete action spaces, demonstrating superhuman performance in games like \cite{alphago}, chess, and shogi \cite{alphazero}. MCTS works by conducting iterative simulations, or rollouts, from the current state and selecting actions based on a tree policy, such as Upper Confidence Bounds for Trees (UCB). As new states are identified, the search tree expands, and each simulation yields an estimate of the expected return. This information is then backpropagated through the tree to refine the value estimates of previous states and actions.

Adapting MCTS to continuous action spaces poses significant challenges. A simple approach would require branching over an infinite range of possible actions, making tree expansion computationally unfeasible. Moreover, the high-dimensional nature of control tasks complicates search efficiency, making real-time applications difficult without substantial optimizations.

\subsection{Beam Search in Planning and RL}

\textbf{Beam search} \cite{beamsearch1, beamsearch2} is a heuristic search method that maintains a set, or \emph{beam}, of $B$ candidate solutions at each step. Each candidate is expanded into multiple successors, and only the top $B$ successors—ranked according to an evaluation function—are retained for further expansion. This approach trades exhaustive exploration for computational efficiency.

In the context of RL and planning:

\begin{itemize}
    \item Beam search can be interpreted as a local look-ahead mechanism, where a set of candidate actions is evaluated at the current time step or over a short horizon. The best candidate is then selected for execution.
    \item Unlike MCTS, which involves a deeper tree-based simulation, beam search provides a more computationally efficient alternative by focusing on evaluating a limited number of high-potential actions.
\end{itemize}

Beam search has been explored in discrete-action RL (e.g., text-based games and combinatorial optimization \cite{beamRL1}), but its application to continuous control remains underexplored due to the difficulty of generating meaningful candidate actions efficiently.

\subsection{Bridging TD3 with Look-Ahead Methods}

Despite the effectiveness of TD3 in continuous control tasks, its reliance on scalar noise for exploration may limit the diversity of actions considered during training. To address this, we propose \emph{MCBS}, which integrates short-horizon planning into TD3 by combining:

\begin{itemize}
    \item \textbf{TD3’s learned value function}, which provides a local, one-step evaluation of action-value estimates.
    \item \textbf{Short-horizon planning}, which introduces a limited but structured form of look-ahead via beam search or MCTS-like rollouts.
\end{itemize}

Instead of relying purely on random noise for exploration, MCBS generates multiple action candidates around the policy’s output and evaluates their expected returns. This evaluation process involves:

\begin{enumerate}
    \item Estimating the Q-value directly for each candidate action using the critic, analogous to beam search.
    \item Conducting short-horizon Monte Carlo rollouts from the current state, gathering rewards, and using the critic to bootstrap the value at the final state. This method incorporates an MCTS-like framework while intentionally keeping the branching factor and depth small to ensure computational efficiency.
\end{enumerate}

By integrating these concepts, MCBS seeks to enhance the equilibrium between exploration and exploitation. The agent gains from a policy that has been fine-tuned through TD3, while also utilizing a constrained approach to short-horizon planning to improve immediate decision-making. In the following section, we will present the algorithmic framework of MCBS and explain how it fits into the conventional TD3 training loop.

\section{Proposed Method: MCBS for TD3}\label{sec:method}

\begin{figure*}[ht]
    \centering
    \includegraphics[width=0.9\textwidth]{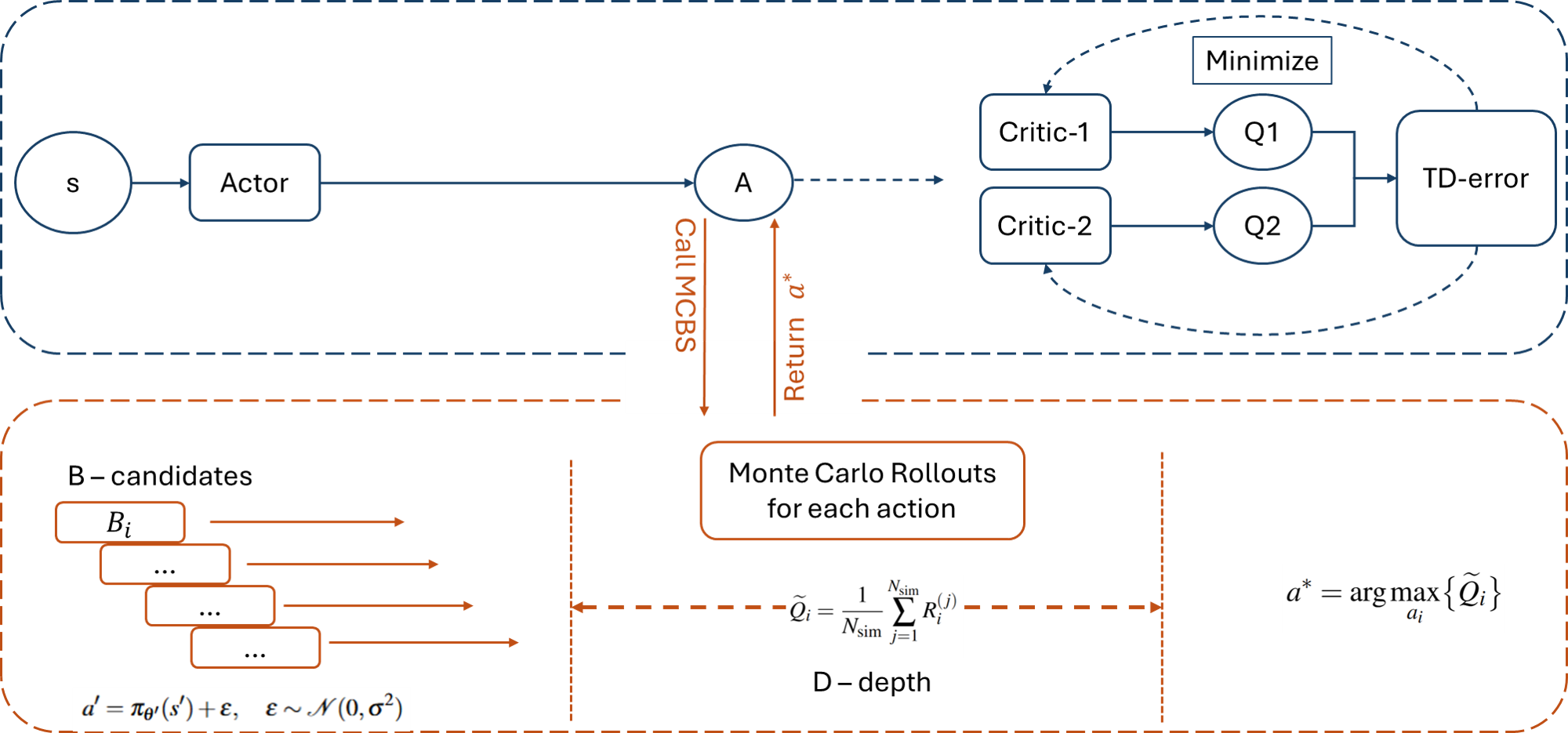}
    \caption{Overview of the Monte Carlo Beam Search enhanced TD3 algorithm. MCBS replaces standard action selection with a structured beam search and Monte Carlo rollouts, improving exploration and policy refinement.}
    \label{fig:mcbs}
\end{figure*}

In this section, we introduce our hybrid approach that augments the standard TD3 algorithm with short-horizon planning through beam search and Monte Carlo rollouts. Our primary objective is to enable more informed action selection in continuous control tasks by leveraging the critic’s learned knowledge alongside local look-ahead simulations. Figure~\ref{fig:mcbs} provides a high-level overview of MCBS-TD3, illustrating how beam search replaces traditional action selection.

\subsection{MCBS Integration}

MCBS enhances TD3 by incorporating a structured look-ahead mechanism to improve action selection. The core steps are as follows:

\begin{enumerate}
    \item \textbf{Policy and Critics (TD3 Core):} We have an actor network $\pi_{\theta}$ that produces continuous actions, along with two critic networks $Q_{\phi_1}$ and $Q_{\phi_2}$ that evaluate the action-value function. These elements adhere to the conventional TD3 framework.
    
    \item \textbf{Action Proposal (Beam Search):} Instead of executing $\pi_{\theta}(s_t)$ directly with added Gaussian noise, we generate a \emph{beam} of $B$ candidate actions near the policy’s output at each state $s_t$. These candidates are obtained by perturbing the policy output with structured noise.

    \item \textbf{Monte Carlo Rollouts:} Each candidate action is assessed using a short-horizon simulation with a depth of $D$, where rewards are gathered throughout the rollout trajectory. At the last step of each rollout, the Q-values estimated by the critics are utilized for bootstrapping to estimate the remaining return.

    \item \textbf{Action Selection:} The candidate action with the highest estimated return is selected for execution. This allows the agent to consider multiple possible actions and refine the immediate decision based on anticipated rewards.

    \item \textbf{Training:} The resulting transition $\bigl(s_t, a_t, r_t, s_{t+1}\bigr)$ is stored in the replay buffer. The actor and critics are updated using the standard TD3 loss functions, with the actor and target networks updated at a delayed frequency to ensure training stability.
\end{enumerate}

\subsection{Advantages of MCBS for Continuous Control}

By assessing various potential actions rather than depending only on random variations for exploration, MCBS facilitates a more organized and insightful search within the action space. The main benefits of this method include:

\begin{itemize}
    \item \textbf{Better Action Selection:} By leveraging both beam search and short-horizon rollouts, the agent can make more informed decisions instead of relying on single-step critic evaluations.
    \item \textbf{Improved Exploration:} Instead of random noise perturbations, MCBS systematically searches for better actions, increasing the likelihood of discovering high-reward behaviors.
    \item \textbf{Enhanced Sample Efficiency:} Short-horizon look-ahead helps the agent exploit the critic’s learned value function, reducing reliance on excessive exploration noise.
    \item \textbf{Robustness to Uncertainty:} Monte Carlo rollouts allow the agent to mitigate potential inaccuracies in Q-value estimation by incorporating multi-step return approximations.
\end{itemize}

By merging the advantages of TD3 with a structured search-based exploration method, MCBS presents a well-founded strategy for overcoming the shortcomings of traditional actor-critic algorithms in continuous control environments. In the next section, we will give a comprehensive overview of the MCBS algorithm and how it fits into the TD3 training process.

\subsection{TD3 Foundation}

Our method builds directly upon the TD3 algorithm \cite{td3paper}. Below, we restate the key elements necessary to contextualize MCBS:

\begin{itemize}
    \item \textbf{Two Critics:} TD3 maintains two independent critic networks, $Q_{\phi_1}(s, a)$ and $Q_{\phi_2}(s, a)$, which are trained via mean-squared error (MSE) regression against a \emph{target value}:
    \begin{equation}\label{eq:td3_target}
        y = r + \gamma \bigl(1 - d\bigr) \min\bigl(Q_{\phi_1'}(s', a'), Q_{\phi_2'}(s', a')\bigr)
    \end{equation}
    where $(s', r, d)$ represents the next state, reward, and done flag, respectively, $\gamma$ is the discount factor, and $(\phi_1', \phi_2')$ denote the target critic parameters updated via Polyak averaging.
    
    \item \textbf{Delayed Policy Updates:} The actor $\pi_{\theta}$ (along with its target network $\pi_{\theta'}$) is updated after every $k$ critic updates (commonly $k=2$). The actor is trained to maximize $Q_{\phi_1}(s, \pi_{\theta}(s))$ over sampled states.
    
    \item \textbf{Target Policy Smoothing:} When computing the target value in \eqref{eq:td3_target}, TD3 adds clipped Gaussian noise to the next action:
    \begin{equation}\label{eq:clip}
        a' = \pi_{\theta'}(s') + \epsilon, \quad \epsilon \sim \mathcal{N}(0, \sigma^2)
    \end{equation}
    where $\epsilon$ is clipped within $[-\delta, \delta]$ to prevent the policy from overfitting to sharp Q-value spikes.
\end{itemize}

These techniques mitigate overestimation bias and enhance stability in continuous control environments. Our contribution builds on this foundation by introducing an improved action selection mechanism that integrates seamlessly into the TD3 framework.

\subsection{MCBS for Action Selection}\label{subsec:mcts}

MCTS is widely used in discrete action spaces, but its direct application to continuous control is computationally infeasible. Instead, we adopt \textbf{shallow rollouts} of depth $D$ to estimate expected returns efficiently.

\subsubsection{Short-Horizon Simulations}

For each candidate action $a_i$, we perform a simulated rollout of length $D$, accumulating rewards and estimating future returns via the critic. The complete procedure is implemented in Algorithm~\ref{alg:mcbs_td3} as the \textsc{ShortHorizon} function.

\subsubsection{Multiple Simulations per Candidate}

To mitigate noise in individual rollouts, each candidate action undergoes $N_{\text{sim}}$ independent simulations. The final Q-value estimate is computed as function the total return obtained from the $j$-th rollout simulation of action:
\begin{equation}
    \widetilde{Q}_i = \frac{1}{N_{\text{sim}}} \sum_{j=1}^{N_{\text{sim}}} R_{i}^{(j)}
\end{equation}

\subsubsection{Action Selection}

The action maximizing the estimated return across all candidates is selected:
\begin{equation}
    a^* = \arg\max_{a_i} \bigl\{\widetilde{Q}_i\bigr\}
\end{equation}
This approach refines TD3’s action selection by considering short-horizon predictions instead of relying solely on noisy exploration.

\subsection{Algorithmic Summary}

Algorithm~\ref{alg:mcbs_td3} details the full MCBS-TD3 framework, where beam search with Monte Carlo rollouts replaces naive action selection in TD3. The remainder of the training process—including buffer storage, critic updates, and policy improvements—follows the standard TD3 pipeline.

\begin{algorithm}[!ht]
\caption{Monte Carlo Beam Search for TD3 (MCBS)}
\label{alg:mcbs_td3}
\begin{algorithmic}[1]
\State \textbf{Input:} Beam width $B$, rollout depth $D$, number of simulations $N_{\text{sim}}$, replay buffer $\mathcal{D}$, actor network $\pi_{\theta}$, critic networks $Q_{\phi_1}, Q_{\phi_2}$, discount factor $\gamma$, update interval $d$, noise scale $\sigma_b$.
\State \textbf{Initialize:} Actor, critics, target networks, and replay buffer $\mathcal{D}$.

\For{each episode}
    \State Initialize state $s_0$.
    \For{$t = 0$ to $T-1$}
        \State \textbf{1) Generate $B$ Candidate Actions}
        \For{$i = 1$ to $B$}
            \State Sample noise $\eta_i \sim \mathcal{N}(0,\sigma_b^2)$
            \State $a_i = \mathrm{clip} \bigl(\pi_{\theta}(s_t) + \eta_i, -A_{\max}, A_{\max} \bigr)$
        \EndFor

        \State \textbf{2) Evaluate Candidates via Rollouts}
        \For{$i = 1$ to $B$}
            \State Compute rollout estimate:
            \[
            \widetilde{Q}_i = \frac{1}{N_{\text{sim}}} \sum_{j=1}^{N_{\text{sim}}} \textsc{ShortHorizon}(s_t, a_i, D)
            \]
        \EndFor

        \State \textbf{3) Select Best Action}
        \[
        a^* = \arg\max_{a_i} \widetilde{Q}_i
        \]

        \State Execute $a^*$, observe $(r_t, s_{t+1}, d_t)$, store in $\mathcal{D}$.

        \If{$d_t$ (terminal flag)} 
            \State \textbf{Terminate episode.}
        \EndIf

        \State \textbf{4) TD3 Update:} Update $Q_{\phi_1}, Q_{\phi_2}$; every $d$ steps, update $\pi_{\theta}$ and target networks.

        \State $s_t \gets s_{t+1}$
    \EndFor
\EndFor

\Function{ShortHorizon}{$s,a,D$}
    \State Initialize $R = 0$, $\alpha = 1.0$, $s_{\text{sim}} = s$, $a_{\text{sim}} = a$.
    \For{$d' = 1$ to $D$}
        \State $(s_{\text{sim}}', r, \text{done}) = \text{EnvStep}(s_{\text{sim}}, a_{\text{sim}})$
        \State Update return: $R \gets R + \alpha \cdot r$, $\alpha \gets \gamma \cdot \alpha$
        \If{\text{done}} 
            \State \Return $R$
        \EndIf
        \State Sample noise $\eta \sim \mathcal{N}(0,\sigma_b^2)$
        \State Update next action:
        \[
        a_{\text{sim}} = \mathrm{clip} \bigl( \pi_{\theta}(s_{\text{sim}}') + \eta, -A_{\max}, A_{\max} \bigr)
        \]
        \State $s_{\text{sim}} \gets s_{\text{sim}}'$
    \EndFor

    \State Bootstrap final return:
    \[
    \widetilde{r} = \min(Q_{\phi_1}(s_{\text{sim}}, a_{\text{sim}}), Q_{\phi_2}(s_{\text{sim}}, a_{\text{sim}}))
    \]
    \State \Return $R + \alpha \cdot \widetilde{r}$
\EndFunction
\end{algorithmic}
\end{algorithm}

\subsection{Computational Complexity Considerations}

In standard TD3, each environment step requires sampling a single action and performing one forward pass through the policy plus one or two forward passes through the critics (for Q-value estimates). By contrast, MCBS demands:

\begin{itemize}
    \item \textbf{Beam Generation:} $B$ forward passes for the actor (though these can be efficiently batched).
    \item \textbf{Monte Carlo Rollouts:} Each candidate $a_i$ incurs $N_{\text{sim}} \times D$ additional environment simulations (or model-based simulations) plus repeated calls to the actor and critics.
\end{itemize}

Hence, the complexity can scale as $O(B \cdot N_{\text{sim}} \cdot D)$ times the cost of naive action selection. However, for moderate beam widths ($B \le 3$--$18$), shallow rollouts ($D \le 3$--$9$), and a small number of simulations ($N_{\text{sim}} \le 5$--$10$), this overhead may be manageable in fast-simulating environments or settings where sample efficiency is paramount.

\subsection{Design Considerations and Extensions}

\begin{itemize}
    \item \textbf{Noise Scales:} The beam noise scale $\sigma_b$ for candidate generation may differ from the TD3 exploration noise $\sigma_{td3}$. Proper tuning of these values can significantly impact exploration efficiency.
    \item \textbf{Adaptive Beam Size:} The beam width $B$ may be adjusted dynamically based on training progress, using a larger beam for broader exploration in early training and a smaller beam for exploitation later.
    \item \textbf{Hierarchical Rollouts:} A potential extension involves hierarchical policies to reduce the effective dimensionality of the rollout space, making deeper look-ahead more computationally feasible.
    \item \textbf{Partial MCTS Tree Expansion:} Instead of fully resampling an action at each rollout step, a partial tree expansion could be used to improve planning accuracy at the cost of additional computation.
    \item \textbf{Model-Based Rollouts:} If a learned or known environment model is available, short-horizon rollouts can be performed within the model rather than the real environment, reducing computational cost while introducing potential model bias.
\end{itemize}

\section{Experiments and Results}\label{sec:experiments}

In this section, we present our experimental setup and empirical evaluation of the proposed MCBS for TD3 (\textbf{MCBS}). We test MCBS on a suite of continuous-control benchmarks and compare it against prominent baselines, including \textbf{TD3}, \textbf{Soft Actor-Critic (SAC)}, \textbf{Advantage Actor-Critic (A2C)}, and \textbf{Proximal Policy Optimization (PPO)}. We leverage pre-trained baseline models from \textbf{Stable-Baselines3 RL Zoo} \cite{rlzoo}, ensuring fair and reproducible comparisons. Additionally, we conduct ablation studies on key MCBS hyperparameters (beam width $B$ and rollout depth $D$), analyzing sample efficiency, final performance, and computational overhead.

\subsection{Experimental Setup}

\subsubsection{Environments}
We evaluate MCBS on standard continuous-control tasks from OpenAI Gymnasium \cite{gym}, implemented via MuJoCo \cite{mujoco1}:

\begin{itemize}
    \item \textbf{HalfCheetah} (\texttt{HalfCheetah-v4}): A 6-DoF planar cheetah robot optimized for high-speed locomotion.
    \item \textbf{Walker2d} (\texttt{Walker-v5}): A two-legged humanoid walker navigating stable locomotion.
    \item \textbf{Swimmer} (\texttt{Swimmer-v5}): A low-dimensional swimming agent optimizing movement in a frictional environment.
\end{itemize}

These environments vary in action dimensionality and complexity, providing diverse benchmarks for evaluating continuous RL algorithms.

\subsubsection{Baselines}
We compare \textbf{MCBS-TD3} to:
\begin{itemize}
    \item \textbf{TD3 (Baseline)} \cite{td3paper}: A deterministic policy gradient method featuring twin Q-functions, delayed policy updates, and target policy smoothing.
    \item \textbf{Soft Actor-Critic (SAC)} \cite{sacpaper}: A maximum-entropy RL algorithm that learns a stochastic policy while optimizing an entropy-regularized critic objective.
    \item \textbf{Advantage Actor-Critic (A2C)} \cite{a2cpaper}: A synchronous policy gradient method that incorporates a learned baseline for variance reduction.
    \item \textbf{Proximal Policy Optimization (PPO)} \cite{ppopaper}: A widely used on-policy RL method that constrains policy updates via a clipped surrogate objective.
\end{itemize}

\subsection{Results}

\subsubsection{Learning Curves}

\begin{figure}[t]
    \centering
    \includegraphics[width=0.48\textwidth]{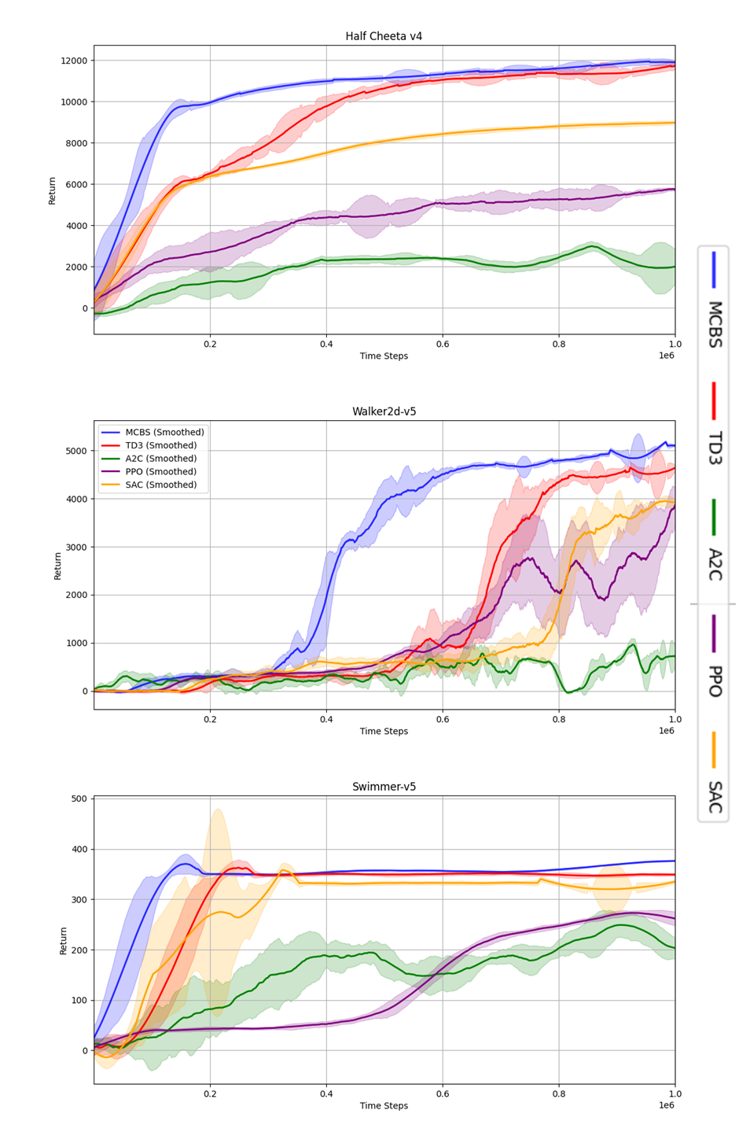}
    \caption{Learning curves for \textbf{HalfCheetah-v4}, \textbf{Walker2D-v5}, and \textbf{Swimmer-v5}, comparing MCBS-TD3, A2C, PPO, TD3, and SAC over 1 million timesteps.}
    \label{fig:learning_curves_mujoco}
\end{figure}

Figure~\ref{fig:learning_curves_mujoco} presents the learning curves for \texttt{HalfCheetah-v4}, \texttt{Walker2d-v5}, and \texttt{Swimmer-v5}, comparing \textbf{MCBS-TD3}, \textbf{TD3}, \textbf{A2C}, \textbf{PPO}, and \textbf{SAC}. MCBS-TD3 consistently demonstrates faster convergence and improved performance relative to other algorithms, particularly in early to mid-training, suggesting enhanced sample efficiency due to short-horizon planning. This reinforces our initial assumption of faster convergence due to larger and more efficient exploration.

\begin{figure}[t]
    \centering
    \includegraphics[width=0.5\textwidth]{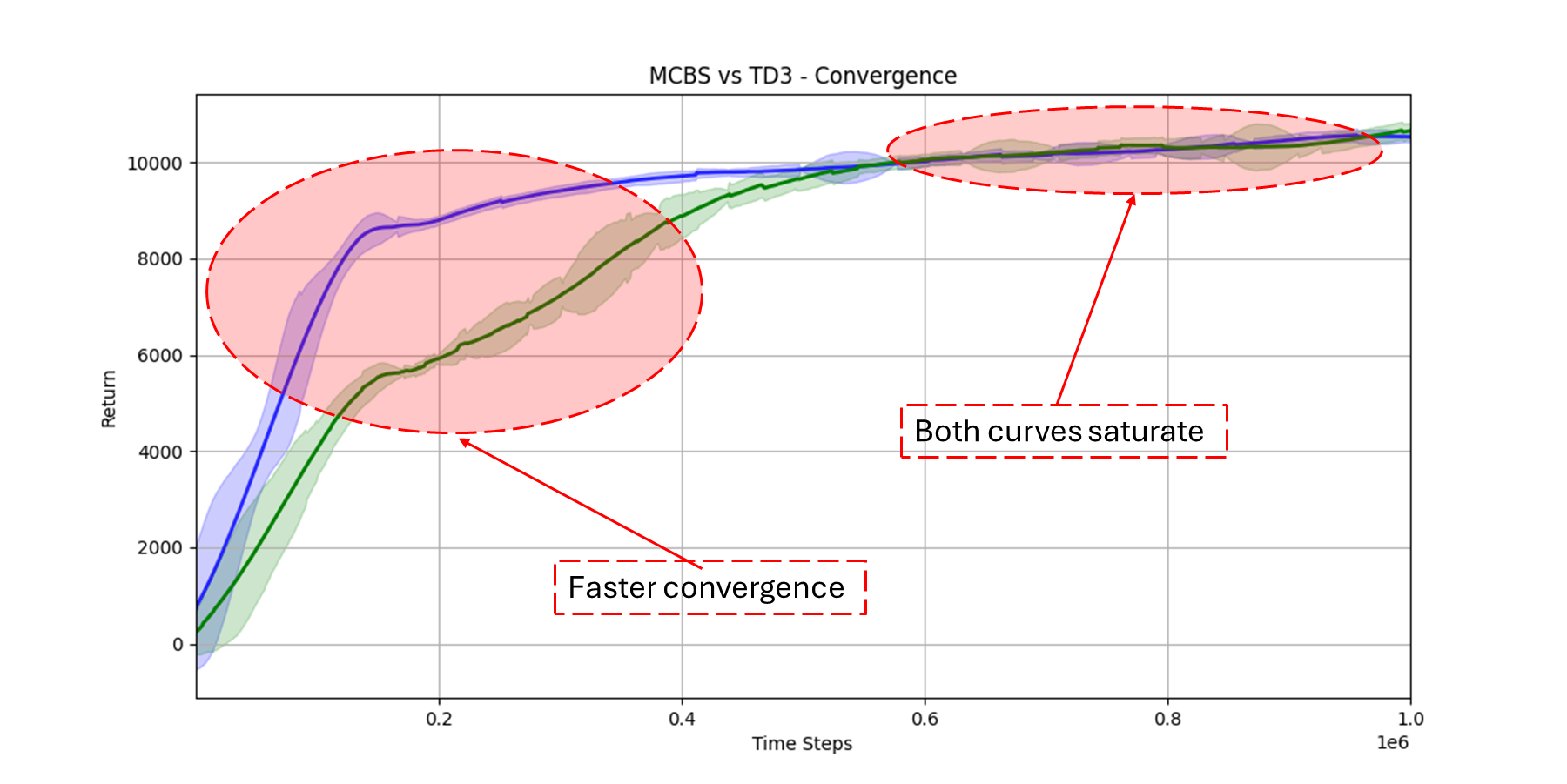}
    \caption{Comparison of MCBS-TD3 and standard TD3 across environments. MCBS significantly accelerates learning early on, but reducing rollout frequency after stabilization is crucial for efficiency.}
    \label{fig:mcbs_vs_td3}
\end{figure}

To further investigate the improvements of MCBS over TD3, Figure~\ref{fig:mcbs_vs_td3} provides a direct comparison. Notably, while MCBS initially accelerates learning, maintaining high-frequency rollouts throughout training can lead to unnecessary computational overhead once the policy stabilizes. To address this, we introduced an adaptive frequency mechanism based on reward saturation, dynamically reducing the rollout frequency as the policy converges. This adjustment significantly improves efficiency while maintaining performance. Our results show that MCBS-TD3 achieves 90\% of the maximum return between 200K and 300K timesteps earlier than vanilla TD3, demonstrating its enhanced sample efficiency and faster convergence.

\subsubsection{Comparison of Final Performance}

Table~\ref{tab:final_performance} summarizes final average returns for TD3, SAC, and MCBS-TD3 across all evaluated environments. MCBS-TD3 consistently achieves higher returns than TD3; it is also notable that SAC remains competitive due to its stochastic exploration.

\subsubsection{Ablation: Beam Width and Depth}

Table~\ref{tab:beam_depth_ablation} reports average returns (at $N_{\text{steps}} = 1$ million) for different beam widths ($B$) and rollout depths ($D$). Key observations:
\begin{itemize}
    \item Increasing $B$ enhances performance up to $B=18$, but further gains diminish due to increased computational cost.
    \item Deeper rollouts ($D \geq 3$) further improve returns, though at the cost of higher runtime.
\end{itemize}

\begin{table}[!ht]
    \centering
    \caption{Final performance comparison (mean $\pm$ std returns) at 1M steps. MCBS-TD3 uses $(B=6, D=3)$ as a representative configuration.}
    \setlength{\tabcolsep}{2.5pt}

    \label{tab:final_performance}
    \begin{tabular}{l|c|c|c}
        \hline
        \textbf{Environment} & \textbf{SAC} & \textbf{TD3} & \textbf{MCBS-TD3} \\
        \hline
        HalfCheetah-v4 & 8155.99 $\pm$ 139.38 & 9006.67 $\pm$ 145.93 & \textbf{10470.92 $\pm$ 260.6}\\
        Walker2d-v5    & 4003.2 $\pm$ 73.28 & 4671.38 $\pm$ 98.81 & \textbf{5123.95 $\pm$ 15.33}\\
        Swimmer-v5     & 333.62 $\pm$ 2.98 & 349.92 $\pm$ 4.04 & \textbf{376.56 $\pm$ 1.45}\\
        \hline
    \end{tabular}
\end{table}

\begin{table}[!ht]
    \centering
    \caption{Final average returns ($\pm$ std) on \texttt{HalfCheetah-v4} for MCBS-TD3 at 1M steps, varying beam width $B$ and rollout depth $D$.}
    \setlength{\tabcolsep}{3pt}
    \label{tab:beam_depth_ablation}
    \begin{tabular}{c|ccc}
        \hline
        \multirow{2}{*}{$B$} & \multicolumn{3}{c}{$D$}\\
        \cline{2-4}
        & 1 & 3 & 6 \\
        \hline
        $1$ (TD3)  & 9006.67 $\pm$ 145.93 & -- & --\\
        $6$        & 9703.14 $\pm$ 210.22 & 10470.92 $\pm$ 268.64 & 10650.74 $\pm$ 280.05\\
        $18$       & 10040.1 $\pm$ 190.78 & 10653.47 $\pm$ 265.88 & 10840.47 $\pm$ 247.58\\
        \hline
    \end{tabular}
\end{table}

\subsection{Discussion of Findings}

Overall, our experiments highlight the effectiveness of MCBS in continuous control. The main conclusions are:
\begin{itemize}
    \item \textbf{Improved Sample Efficiency:} MCBS-TD3 consistently surpasses TD3 in both learning speed and final returns. Its structured look-ahead exploration allows for improved utilization of the critic’s learned estimates.
    
    \item \textbf{Robustness vs. SAC:} While SAC excels in certain tasks thanks to its stochastic policy, MCBS-TD3 stays competitive, showing that short-horizon planning can enhance actor-critic methods beyond just basic noise-based exploration.

    \item \textbf{Adapting Rollout Frequency:} Keeping a high MCBS frequency during training can lead to unnecessary computational load once the critic has stabilized. Gradually decreasing $B$ or $D$ after the initial training phase can enhance efficiency while still maintaining performance.
    
    \item \textbf{Scalability Considerations:} In high-dimensional tasks, reducing $B$ or $D$, using partial expansions, or leveraging model-based rollouts may improve efficiency. Future work could explore adaptive beam widths or hierarchical policies to further optimize compute usage.

    \item \textbf{Applicability:} While our study is based solely on simulations, MCBS utilizes short-horizon rollouts (\(D\leq6 \)) and adds a maximum of \(B\cdot D \) additional forward passes for each step in the environment. This means that the cost per step increases linearly with both the beam width \(B\) and the rollout depth \(D\), allowing the control loop to maintain real-time performance on a desktop GPU. A comprehensive sim-to-real evaluation will be addressed in future research.
    
\end{itemize}

\section{Conclusion}

This work presents Monte Carlo Beam Search (MCBS) as an enhancement to TD3 for continuous control tasks. By incorporating structured exploration through beam search and short-horizon Monte Carlo rollouts, MCBS greatly enhances sample efficiency and learning stability. Our experiments showed that MCBS-TD3 converges at a higher rate than the standard TD3, demonstrating its effectiveness in exploration. Furthermore, the implementation of an adaptive rollout frequency mechanism based on reward saturation minimized unnecessary computational overhead while preserving performance.

Future work will concentrate on enhancing MCBS for practical robotics applications, especially in trajectory optimization for physical robotic systems. Additional improvements will include dynamically optimizing beam width and rollout depth, incorporating model-based planning, and testing the algorithm in hardware environments where computational efficiency and safety constraints are paramount. In addition, MCBS could be paired with real-time perception-safety modules such as the VisionGPT framework for zero-shot anomaly detection in navigation scenes \cite{visiongpt} to create a unified perception-and-control stack for autonomous robots.

\bibliographystyle{IEEEtran}

\end{document}